\documentclass[manuscript,screen]{acmart}

\usepackage{tabularx}

\begin{document}

\title{Political Text Scaling Meets Computational Semantics
%
}

\author{Federico Nanni}
\email{fnanni@turing.ac.uk}
\orcid{0000-0003-2484-4331}

\author{Goran Glavaš}
\email{goran@informatik.uni-mannheim.de}
\orcid{0000-0002-1301-6314}

\author{Ines Rehbein}
\email{ines@informatik.uni-mannheim.de}
\orcid{0000-0002-9615-6389}

\author{Simone Paolo Ponzetto}
\email{simone@informatik.uni-mannheim.de}
\orcid{0000-0001-7484-2049}

\author{Heiner Stuckenschmidt}
\email{heiner@informatik.uni-mannheim.de}
\orcid{0000-0002-0209-3859}
\affiliation{%
  \institution{Data and Web Science Group, University of Mannheim}
  \streetaddress{B6 26}
  \city{Mannheim}
  \country{Germany}
  \postcode{68159}
}

\renewcommand{\shortauthors}{Nanni et al.}

\begin{abstract}
During the last fifteen years, automatic text scaling has become one of the key tools of the Text as Data community in political science. Prominent text scaling algorithms, however, rely on the assumption that latent positions can be captured just by leveraging the information about word frequencies in documents under study. We challenge this traditional view and present a new, semantically aware text scaling algorithm, \textit{SemScale}, which combines recent developments in the area of computational linguistics with unsupervised graph-based clustering.
We conduct an extensive quantitative analysis over a collection of speeches from the European Parliament in five different languages and from two different legislative terms, and show that a scaling approach relying on semantic document representations is often better at capturing known underlying political dimensions than the established frequency-based (i.e., symbolic) scaling method. We further validate our findings through a series of experiments focused on text preprocessing and feature selection, document representation, scaling of party manifestos, and a supervised extension of our algorithm. To catalyze further research on this new branch of text scaling methods, we release a Python implementation of \textit{SemScale} with all included data sets and evaluation procedures. 
\end{abstract}





\maketitle

\section{Introduction}
\label{sec:intro}

In recent years, automatic \textit{text scaling} has become one of the key tools of the Text as Data community in political science.  
A variety of models have been developed for text scaling and have expanded the scope and focus of political text analyses, thus sustaining the growth of the Text as Data community \citep[\textit{inter alia}]{laver_extracting_2003,slapin_scaling_2008,Lowe2011ScalingTexts}. Text scaling approaches, such as the widely popular Wordscores
\citep{laver_extracting_2003} and Wordfish \citep{slapin_scaling_2008} algorithms, offer the possibility of identifying latent positions of political actors directly from textual evidence produced by those actors, such as party manifestos or political speeches.
Such positions, depending on the type of data and the context of the study, have been interpreted as capturing political preferences such as a Right-Left scale of political ideology or different attitudes towards the European integration process (see, for instance, \citet{laver_extracting_2003} and \citet{Proksch2010PositionSpeeches}).

The Wordfish algorithm has since been applied in many text-analytic studies in political science. 
However, while text scaling methods have shown their potential to interpret textual content directly as a form of political data that can be analysed automatically, it is important to notice that they suffer from a major limitation. 
Namely, they treat textual data in a \textit{symbolic} fashion, i.e., they represent documents simply as bags of words and assign them (explicitly or implicitly) position scores based on the words they contain. This means that the amount of lexical overlap between two texts directly determines the extent of their positional (dis)agreement. This gives rise to two types of errors in position estimation that methods based on lexical overlap are prone to: 
\begin{enumerate}
    \item Texts that convey similar meaning and express a similar political position but overlap only in very few words (e.g., \textit{``\dots homophobic outbursts should have no place in modern German society.''} and \textit{``\dots anti-gay propaganda needs to be prevented.''}) will end up being assigned very different position scores.
    \item Texts that convey different or opposing political positions but have a significant word overlap (e.g., \textit{``Migrants are responsible for the increased crime rates.''} vs. \textit{``Migrants are responsible for fewer crimes than domicile population.''}) will end up being assigned similar position scores.
\end{enumerate}

In other words, the unlimited expressiveness of natural language not only allows us to express similar political positions in very different ways and without any word overlap, but also to lexicalize very different political positions, based on similar words. 
In this work, we address the first issue and propose a scaling approach that remedies for the above-mentioned limitation of existing scaling methods by considering \textit{semantic} representations of words in a text. In general, \textit{semantic word representations} are computational representations (e.g., vectors) that have the following property: words with similar meaning (e.g., \textit{``homophobic''} and \textit{``anti-gay''}) have similar representations; conversely, words with a different meaning (e.g., \textit{``propaganda''} and \textit{``cheese''}) should have dissimilar computational representations.

Our semantic scaling algorithm, dubbed SemScale, is based on semantic representations of words instead of the words itself, thus leveraging recent developments in the area of computational linguistics where methods for inducing robust algebraic representations of word meaning have been proposed \citep[\textit{inter alia}]{Mikolov,pennington2014glove,Bojanowski2017EnrichingInformation}. By relying on semantic rather than symbolic representations of text, SemScale can distinguish between words with different meaning and phrases with similar or related meaning (e.g., that \textit{``homophobic outbursts''} has a similar meaning as \textit{``anti-gay propaganda''}) and can make use of such semantic similarities to produce the scaling scores. Additionally, SemScale is a fully deterministic algorithm, which helps to address issues of consistency and reproducibility of results obtained via text mining approaches.

To assess the benefits of our new, unsupervised\footnote{\textit{Unsupervised} refers to the fact that the algorithm does not require any text to be assigned position scores by human annotators. An algorithm that requires some texts to be annotated with position scores (assigned by human annotators), in order to be able to predict the scores for other texts, is, in contrast, a \textit{supervised} algorithm. Wordscores \citep{laver_extracting_2003} is an example of a supervised text scaling algorithm.} approach to text scaling, we present an extensive empirical comparison of SemScale with the most widely adopted unsupervised text scaling algorithm, Wordfish, which, by operating merely on the basis of word frequencies, is unaware of word meaning. 
We assess the robustness of our results across different languages and time periods. To do so, we created a benchmarking data set for text scaling from the European Parliament website which comprises all speeches given by members of the European Parliament (MEP) in one of the following five languages, English, German, French, Italian, and Spanish, and their official translations to all of the other languages during the 5th and 6th legislative terms.  

Our data set creation builds on the work of \citet{Proksch2010PositionSpeeches} and can be seen as an extension of the data set used in their work for testing the robustness of Wordfish, which covered only the speeches produced in or translated into English, German, and French and only during the 5th legislative term.
Our work is thus able to shed more light on the robustness of different text scaling approaches and their sensitivity to different preprocessing methods, the choice of text representation and topical changes in the input data across time.

The main contribution of this study is a novel unsupervised algorithm for text scaling, based on semantic text representations. We demonstrate empirically that our method outperforms the widely adopted Wordfish algorithm in terms of identifying party positions on European integration. 
In order to stimulate further research and collaborations on semantically-aware text scaling, we release (as supplementary material to this work) \citep{nanni2019app}: (i) the multi-language data set employed in this study (in its original form and after each preprocessing step), (ii) all scaling results (i.e., individual party positions) obtained in our work and (iii) an offline Python implementation of SemScale (a command-line tool).\footnote{\url{https://github.com/umanlp/SemScale}}
 
In addition to this, we provide a series of validation experiments addressing central points of the current debate on text scaling algorithms. First of all, an often criticised aspect of existing scaling methods is their inability to decipher which (if any) underlying policy dimension is captured by the produced position scores (cf. for instance, the critiques raised by \citet{Budge2007DoSeries,Budge2007Missingresponse} concerning Wordscores or the recommendation made by \citet{Proksch2010PositionSpeeches} with respect to filtering ideological from non-ideological statements prior to applying Wordfish). In addition to the above criticism, \citet{Denny2018TextIt} have recently questioned the robustness of Wordfish, demonstrating that it is highly sensitive to even small changes in the input text, such as the removal of punctuation or stop words, which should have no effect on the overall political message, i.e., position. To address this issue, we examine the robustness and stability of text scaling results for different lexical and semantic representations of the input texts.
We are particularly interested in a better understanding of the extent to which \emph{a)} known policy positions are captured by specific linguistic traits, such as specific parts of speech or named entities, in contrast to using the entire texts, and \emph{b)} whether this is further emphasised by our newly proposed scaling approach, which also captures word meaning (and not just the frequency of words in different texts).

Our second validation experiment is aimed to shed light on the contribution of dense semantic text representations for SemScale, in comparison to other types of text representations. In particular, we conduct several comparisons where we employ our newly proposed scaling algorithm and substitute word-embeddings (which we use as textual representations) with: \emph{a)} term frequency–inverse document frequency (TF-IDF) vectors and \emph{b)} newly proposed party vectors \citep{rheault2019word}. The first comparison will unveil whether semantic vectors are in fact necessary to better determine positions or whether the core contribution of our approach comes from its new graph-based scaling algorithm. The second comparison will reveal whether directly inducing document-level representation vectors \citep{le2014distributed} more precisely captures party positions, compared to representing each document as an aggregation of its word embeddings.

The third validation experiment compares the performance of SemScale and Wordfish on different types of texts, to determine the consistency of our results across different textual sources. 
For this we compare results obtained on speeches from the European Parliament with the ones obtained on party manifestos from five different countries. We study scaling performance (i) when positioning only manifestos from the same elections and (ii) when positioning manifestos from multiple elections on a single scale.

Our next validation experiment investigates the impact of different graph-based clustering algorithms on the results and reports scaling scores for two different, well-known algorithms (i.e., the {\em harmonic function label propagation} algorithm (HFLP) and the {\em PageRank} algorithm).

We then focus on the core ingredient of SemScale, the dense semantic representations, and explore whether we can improve Wordfish by adding semantic information on word similarity to the input. For that, instead of providing Wordfish with a document frequency matrix based on word frequencies, we compute the frequency matrix by grouping similar words together, based on the cosine similarity of their distributional semantic representations, and counting frequencies of word groups in the documents. This experiment is meant to disentangle the impact of the semantic representations from that of the algorithm used for text scaling that we studied in the previous experiment.

In our final validation experiment, we compare SemScale with another widely used scaling algorithm, Wordscores, a \textit{supervised} scaling approach that positions so-called virgin texts (i.e., texts of unknown positions) on the basis of known positions of given reference texts. To allow for a fair comparison, we extend SemScale to take two "pivot texts" as supervised inputs, thereby determining the extremes of the scale. While SemScale is primarily designed as an unsupervised algorithm to operate in low-resource settings in which we do not expect to find human annotations of position scores, here we investigate its performance and potential limitations in (weakly) supervised settings where a small number of texts with assigned positions actually exist. 
The structure of the article is as follows. 
In Section \ref{sec:previous} ({\sc Previous Work on Political Text Scaling}), we discuss related work on political text scaling and describe the well known  Wordfish \citep{slapin_scaling_2008} and Wordscores \citep{laver_extracting_2003}  algorithms.
Section \ref{sec:semscale} (\textsc{SemScale -- A Semantic Model for Political Text Scaling}) provides a detailed description of SemScale, our newly proposed scaling method that exploits semantic representations of words and texts. 
Section \ref{sec:eval} (\textsc{Quantitative Evaluation}) presents the data and setup used in our experiments and reports our results.   
Additional validation experiments are presented in Section \ref{sec:validation} (\textsc{Validation Experiments}) where we investigate the impact of different types of features and text representations on the results and test the robustness of different scaling algorithms in various settings.
We conclude by discussing our findings and the implications they might have for fostering further research on semantically-aware analyses of political texts.

\section{Previous Work on Political Text Scaling}
\label{sec:previous}

Positioning political actors along predefined dimensions in ideological spaces has been an important foundation of research in the area of political science. One of the most famous examples is the Chapel Hill Expert Survey that relies on expert ratings for estimating party positions on topics like European integration, political ideology and policy issues for national parties in a variety of European countries \cite{Polk2017ExplainingData,bakker:etal:2020}. 
Another example is the Global Party Survey (GPS) \cite{norris:2020}, which also relies on expert surveys in order to position parties on an ideological scale and to obtain measures of populism for parties across the world.
A related effort is the Global Populism Database \cite{hawkins:etal:2019} which positions more than 200 international political leaders according to their degree of latent populist content, based on the speeches they delivered. Those speeches have been subject to holistic grading by human experts, in order to measure latent aspects of populism in the texts. 

Those are only a few examples illustrating the importance of spatial models of politics where political attitudes and preferences such as Left-Right ideology or the degree of populism  are conceptualised as positions in latent space. 
However, obtaining information on political ideology or other variables of interest for subjects across the world and over a long period of time is extremely costly in terms of time and manpower.
Therefore, many recent studies have been focused on inferring latent political positions directly from the texts produced by political actors \cite{poole:rosenthal:1985,martin:quinn:2002,Proksch2010PositionSpeeches,Hjorth2015ComputersPositions,Laver2018ExtractingData,egerod:klemmsen:2019}, to bypass time-consuming data acquisition and manual coding.
Most of these works on political text scaling can be divided into two branches, supervised and unsupervised approaches to text scaling. 
Below we present the two most prominent algorithms for each branch and discuss their merits and drawbacks.

\subsection{Supervised approaches to political text scaling: Wordscores}

One of the most widely adopted \textit{supervised} text scaling algorithms is Wordscores. 
Introduced by \citet{laver_extracting_2003}, Wordscores is built around the assumption that word frequency information from ``reference'' texts, for which the position scores for the dimension of interest have been provided by human annotators, can be used to make predictions for new texts for which the positions are unknown. 
It is important to note that the notion of \textit{supervision} used here is different from supervised machine learning which relies on a large number of training instances in order to learn generalisations for new data, while Wordscores requires only a few data points to define the dimension of interest.

Let us assume that we have a number of reference texts $r$ with known positions on the scaling dimension of interest and a data set with new texts that we want to position. The first step consists of iterating over each word $w$ in the reference texts $r$ and determining the position of $w$, based on its frequency in the reference texts.
In the second step, we now iterate over the words in the new, unlabelled texts $u$, ignoring all words that we haven't seen in the reference texts. We can then compare the positions of the words in $r$ to their positions in $u$ and use this information to assign a score to the new documents.
The wordscores $S_{wd}$ for each word $w$ on dimension $d$ are computed as follows:
\begin{equation*}
 S_{wd} = \sum_r P_{wr} * A_{rd}
\end{equation*}
where $P_{wr}$ is the probability of word $w$ occurring in document $r$ and $A_{rd}$ is the known position of the document $r$ on dimension $d$.
Once we have computed the wordscores $S_{wd}$ for each word in the reference texts $r$, we can use them to infer the position of the new, unknown texts $S_{ud}$ by comparing the wordscores to the word frequencies in the new texts:
\begin{equation*}
 S_{ud} = \sum_w F_{wu} * S_{wd}
\end{equation*}
where $F_{wu}$ is the frequency of each word $w$ in the new, unlabelled texts $u$.
As pointed out by Egerod and Klemmsen \cite{egerod:klemmsen:2019}, Wordscores is based on a number of unrealistic assumptions, including the assumption that each word in the texts is equally informative and that the words in the new, unseen documents come from the same distribution as the ones in the reference texts.
While these assumptions are not met in real life, they also point out that they can at least help to mitigate bias when used to guide the selection of the reference texts.
Since its first introduction, a number of extensions to Wordscores have been presented that address some of these issues \cite{lowe:2008,martin:vannberg:2008,Perry2017ScalingModel} (see \cite{egerod:klemmsen:2019} for an overview).

\subsection{Unsupervised approaches to political text scaling: Wordfish}

The most known unsupervised algorithm for political text scaling is Wordfish \cite{slapin_scaling_2008,Proksch2010PositionSpeeches}. 
One crucial difference between the \textit{unsupervised} Wordfish and Wordscores is that, while for Wordscores the scaling dimension is given by means of a small number of reference texts and their scores on the respective dimension, for Wordfish the dimension of interest is unknown and has to be inferred. While this avoids the problem of inserting bias, for example by means of poorly chosen reference texts, it also makes the learning problem much harder.

Wordfish assumes that the words in the documents follow a Poisson distribution. 
More specifically, Wordfish is a variant of a Poisson ideal point model where, given a collection of documents, the $j$-th vocabulary word's frequency in the $i$-th document, $W\textsubscript{ij}$s is drawn from a Poisson
distribution with rate $\lambda\textsubscript{ij}$, which is modeled considering the document length ($\alpha\textsubscript{i})$, the token-frequency ($\psi\textsubscript{j}$), the level to which a token identifies the direction of the underlying ideological space ($\beta\textsubscript{j}$), and the underlying position of the document ($\theta\textsubscript{i}$):
\begin{equation*}
    \lambda\textsubscript{ij} = exp(\alpha\textsubscript{i} + \psi\textsubscript{j} + \beta\textsubscript{j} \times \theta\textsubscript{i}).
\end{equation*}

\subsection{Strengths and weaknesses}

In many cases the more explorative unsupervised approach of learning the scaling dimension jointly with the document positions might seem attractive, as this might uncover latent dimensions hidden in the data, while the supervised approach is restricted to use the information that we predefine as the dimension of interest.
However, not providing the model with enough guidance might also cause the model to pick up on irrelevant aspects such as topic distinctions \cite{Lowe2013ValidatingBenchmark} instead of ideological positions. 
Egerod and Klemmsen \cite{egerod:klemmsen:2019} therefore stress the point that automated political text scaling should only be used to support human experts' analyses and not replace them (also see \cite{grimmer_text_2013} for a critical survey of automated content analysis methods for political text analysis).

A major strength of models relying on word frequencies, such as Wordscores and Wordfish, is that 
they are directly applicable in any language precisely because they do not explicitly model semantics but adopt word frequency rates as a (often successful) proxy to document semantics. 
This, however, also comes with a major drawback, as already pointed out in Section \ref{sec:intro}.
The symbolic nature of the representations used in both models, Wordscores and Wordfish, consider semantically similar words such as {\em company} and {\em firm} as equally distinct as very dissimilar words like {\em company} and {\em sunflower}, and thus rely on a substantial overlap in vocabulary between texts, precisely because the models are not able to generalize and identify similar meaning across word forms.
This is also the motivation for our novel approach to text scaling, described in the next section, which positions texts based on the {\em semantics} of the words in the documents, instead of the words itself. 

\section{SemScale -- a Semantic Model for Political Text Scaling}
\label{sec:semscale}

\begin{figure}
\begin{center}
\includegraphics[scale=1,bb=0 0 421 442]{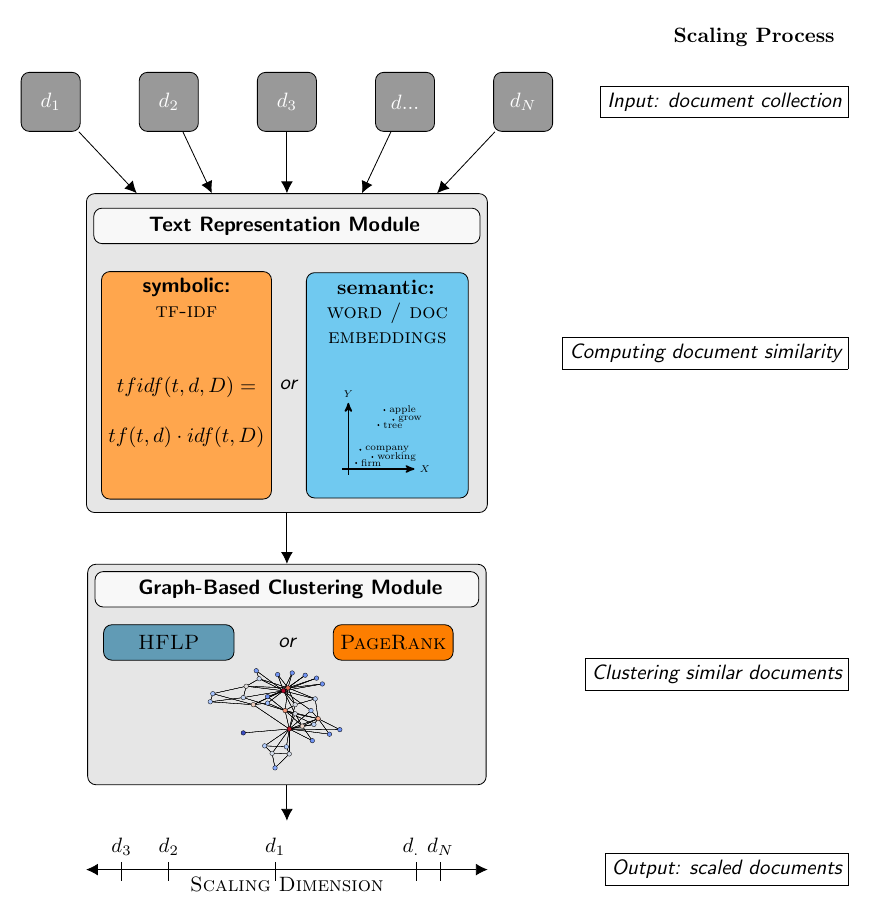}
\end{center}
\caption{Overview of the different components of our modular approach to text scaling.}\label{fig:semscale} 
\end{figure}

Existing scaling algorithms like Wordfish and Wordscores model words as discrete symbols and consider only their (weighted) frequency; as opposed to this, modern research in computational linguistics primarily represents words as numeric vectors sampled from a continuous multi-dimensional vector space. The research area of {\em distributional semantics}, in particular, builds upon the assumption that the meaning of a word can be derived by looking at the contexts in which it is used or, as \citet{Firth1957} puts it, that ``a word is characterised by the company it keeps''. For instance, if we consider the sentence ``The members of \emph{+Europa} voted against the proposal'', even without knowing what \emph{+Europa} is, we can infer from the context that it is probably a political entity.

The ability to precisely capture the meaning of words by representing them as points in a multi-dimensional semantic vector space, i.e., by representing words with  so-called \textit{word embeddings} \citep{Mikolov}, is arguably one of the most relevant achievements of computational linguistics in the last few decades. Among other things, \textit{word embeddings} can be used to detect particular semantic relations that hold between words (e.g., hypernymy, synonymy, antonymy, or meronymy) \citep{glavavs@emnlp2017,glavavs2018discriminating} or between entities (e.g., ``being capital of'', ``being president of'') \citep{nickel2016review,joulin2017fast}.

In this work, we examine the potential of distributional semantics for obtaining  vector representations of texts for unsupervised political text scaling.\footnote{Word embeddings have already been studied and employed in political science analyses, e.g., in \citet{gurciullo2017detecting},  \citet{Spirling2019} and \citet{rheault2019word}. In this work, we test their potential for capturing political positions in a latent space.} 
As Wordfish cannot handle distributional representations of texts but requires symbolic representations (i.e., words) as input, we developed SemScale, a new algorithm for political text scaling, based on distributional semantic representations and graph-based clustering, that we now describe in more detail.

SemScale has two major components, a) the {\em text representation} module, and b) the {\em graph-based text scaling} algorithm (see Figure \ref{fig:semscale}).\footnote{An earlier version of the algorithm, with further technical details and an extension for cross-lingual text scaling, is described in \citet{Glavas2017UnsupervisedTexts}.}  
This means that, unlike Wordfish, our core graph-based scaling algorithm is detached from the document representation and can work with both, symbolic and distributional semantic text representations (as demonstrated in Section \ref{sec:validation}, \textsc{Validation Experiments}). 
This makes the model highly flexible and allows us to compute the similarity between two documents as the similarity between their symbolic term-frequency--index-document-frequency (\textit{tf}-\textit{idf}, cf.~the definition in the next section) vectors, but also to compute the similarity between two texts based on dense semantic representations, such as word embedding vectors.

\subsection{Text representation} 
We now describe the default version of SemScale, in which we represent texts as averages of word embeddings \citep{Bojanowski2017EnrichingInformation,glavavs2018resource}. In Section \textsc{Validation Experiments}, we demonstrate how the SemScale's graph-based scaling algorithm can be coupled with (2) traditional symbolic tf-idf text representations \citep{salton1988term} and (2) document embedding representations \citep{le2014distributed}. \\

We start by representing each document under study by its respective distributional semantic vector, built by aggregating the embeddings of the words in the document as follows: Let $T$ be the bag of words of a political text, i.e., the set of all words that appear in that text, and let $e(w)$ be the word embedding of some word $w$. We then compute the embedding vector of the whole text, $e(T)$, by computing the weighted average of the embeddings of all words in $T$: 
\begin{equation*}
e(T) = \frac{1}{|T|}\sum_{w \in T} \text{tf-idf}(w) \cdot e(w) .
\end{equation*}
$\text{Tf-idf}(w)$ stands for the \textit{term frequency-inverse document frequency} score for  word $w$ and document $T$ and is used as the weight with which we multiply the embedding vector $e(w)$ of the word $w$. The $\text{tf-idf}$ score of the word $w$ for the text $T$ is the product of two scores. The first one is the term frequency score (TF) which captures how often the word appears in the document and the second term is the inverse document frequency score (IDF), which is inversely proportional to the number of other texts in the collection that contain the word $w$.\footnote{The intuition behind the \textit{tf-idf} weighting scheme is that the word contributes more to the overall meaning of the text the more frequently it appears in the document (TF component) and the less common it is, i.e., the lower the number of other texts that contain that same word is (IDF component).} Precisely, we compute the TF score for a word \textit{w} and text document \textit{T} as follows: 
\begin{equation*}
    TF(w, T) = \frac{\mathit{freq}(w, T)}{\max_{w'}{\mathit{freq}(w', T)}}
\end{equation*}
where $\mathit{freq}(w, T)$ is the raw frequency of occurrence of $w$ in $T$, normalized by the maximal frequency with which any word ($w'$) appears in $T$. The IDF is computed instead for each word $w$ as follows: 
\begin{equation*}
    IDF(w) = \ln\frac{|D|}{|\{T \in D: w \in T\}|}
\end{equation*}
where $D$ is the collection of textual documents (and $|D|$ is the number of documents in the collection) and $\{T \in D: w \in T\}$ is the subset of the documents in the collection $D$ that contain the word $w$. 

Then, let $T_1$, $T_2$, $\dots$, $T_N$ be the collection of $N$ political texts which we want to scale, with their corresponding distributional semantic vectors $e(T_1)$, $e(T_2)$, $\dots$, $e(T_N)$, computed from word embeddings as described above. We can then measure the semantic similarity between 
any two texts $T_i$ and $T_j$ by comparing their respective embeddings, i.e., by comparing $e(T_i)$ with $e(T_j)$. Following common practice with respect to vector-space text representations, we measure the semantic similarity between two texts $T_i$ and $T_j$ as the cosine of the angle that their respective embedding vectors enclose:
\begin{equation*}
    \mathit{sim}(T_i, T_j) = \frac{e(T_i) \cdot e(T_j)}{\lVert e(T_i) \rVert \cdot \lVert e(T_j) \rVert}
\end{equation*}
where $e(T_i) \cdot e(T_j)$ is the dot product between vectors $e(T_i)$ and $e(T_j)$ and $\lVert e(T) \rVert$ denotes the Euclidean norm of the vector $e(T)$. By computing the above similarity for every possible pair of texts in our collection,\footnote{In a collection of $N$ texts there are $\frac{N(N-1)}{2}$ different text pairs, i.e., we need to compute $\frac{N(N-1)}{2}$ similarity scores.} we give rise to a fully-connected weighted graph\footnote{A fully-connected weighted graph is a graph in which there is an edge between every two vertices and there is a numeric weight assigned to each edge.} which we call the \textit{similarity graph}. The vertices in the similarity graph denote individual texts in our text collection (i.e., vertex $V_i$ corresponds to the text $T_i$), whereas the weights of the edges denote how semantically similar the two texts are (i.e., the weight of the edge between vertices $V_i$ and $V_j$ is $w_{ij} = \mathit{sim}(T_i, T_j)$). Again, note that, while in the default variant of SemScale we use the cosine similarity between aggregated embedding vectors $e(T_i)$ and $e(T_j)$, one can set $\mathit{sim}(T_i, T_j)$ to be any other function that measures some type of similarity between texts in order to induce the similarity graph. We empirically investigate two other similarity functions in Section \textsc{Validation Experiments}. Our graph-based scaling algorithm that we describe next is completely agnostic to how the similarity scores (i.e., weights of the edges of the similarity graph) have been computed.\\   

\noindent \textbf{Graph-based scaling}. The graph-based scaling algorithm aims to assign a position score to each vertex $V_i$ in the graph, by taking into account the weights of the edges that connect that vertex with all other vertices, that is, by considering the semantic similarity of the corresponding text $T_i$ with all other texts in the text collection $D$. We start from an intuitive assumption that a pair of least similar (i.e., most dissimilar) texts corresponds to extreme positions in the position spectrum. 
In other words, among all possible pairs of texts ($T_i$, $T_j$), we identify those two that have the lowest mutual semantic similarity (i.e., lowest $\mathit{sim}(T_i, T_j)$) and assume that one of them is on one end of the position spectrum, whereas the other is on the opposite end; positions of all other texts are assumed to lay somewhere in between these two extremes. We name these two most dissimilar texts \textit{pivot texts} and assign an initial position score of $1$ to one of them and $-1$ to the other. We next propagate the position scores assigned to the pivot texts to all the other text (which are still without a position score), using the structure and the weights of the similarity graph as the backbone for score propagation. Namely, we employ so-called \textit{harmonic function label propagation} (HFLP) algorithm, proposed by \citet{zhu2003semi} -- a commonly used algorithm for graph-based semi-supervised learning -- to propagate position scores from the two pivot texts to other, non-pivot texts. 
Let $G = (V, E)$ be our similarity graph and $\mathbf{W}$ its weighted adjacency matrix. Let $\mathbf{D}$ be the diagonal matrix with weighted degrees of graph's vertices as diagonal elements, i.e., $D_{ii} = \sum_{j \in |V|}{w_{ij}}$, where $w_{ij}$ is the weight of the edge between vertices $i$ and $j$. Then $\mathbf{L} = \mathbf{D} - \mathbf{W}$ is the unnormalized Laplacian of the graph $G$, a matrix representation of the graph $G$ which can be used to detect many useful properties of $G$. Assuming that the labeled vertices -- the vertices to which we have assigned a position score, i.e., the two vertices corresponding to pivot texts -- are ordered before the unlabeled ones (vertices corresponding to all other texts in our collection), the Laplacian matrix $\mathbf{L}$ of the similarity graph $G$ can be partitioned as follows:                     
\begin{equation*}
\mathbf{L} = \begin{pmatrix}
\mathbf{L_{ll}} & \mathbf{L_{lu}} \\
\mathbf{L_{ul}} & \mathbf{L_{uu}}
\end{pmatrix}
\end{equation*}
The vector containing the scores of the unlabeled vertices (which are vertices corresponding to all but the two pivot texts), capturing the position scores of the non-pivot texts, is then computed as: 
\begin{equation*}
\mathbf{f_{u}} = \mathbf{-L_{uu}^{-1}}\mathbf{L_{ul}}\mathbf{y_l}
\end{equation*}
where $\mathbf{y_l}$ is the vector of scores of labeled vertices, in our case the vector with the scores of pivot vertices, $\mathbf{y_l} = [1, -1]^{T}$. This way, by propagating the position scores from pivot vertices to all other vertices through exploitation of the structure of the similarity graph $G$, we obtain the position scores for all texts in our text collection.\\ 

As Wordfish and SemScale are both unsupervised scaling algorithms, we will first focus on a comparison between the two algorithms (Section \ref{sec:eval}, \textsc{Quantitative Evaluation}). 
It is worth mentioning that, same as Wordfish, SemScale produces a spectrum of position scores but cannot tell the orientation of the scale. For example, given the left-to-right ideological scaling, we do not know whether the leftmost point on the scale produced by SemScale corresponds to the political party which is most to the left in the political spectrum or to the political party which is most to the right. 

SemScale is a fully deterministic algorithm, assuming a fixed collection of pretrained word embeddings. In other words, when using the same pretrained word embeddings, SemScale will always produce the same output (i.e., same positions for texts) given the same input (the same collection of texts). In contrast, various Wordfish implementations all obtain model parameters via stochastic optimisation methods, which may lead to somewhat different results being produced by multiple runs on the same data input.

In summary, our new scaling algorithm provides a flexible architecture which allows us to plug in different types of text representations and to test their impact on political text scaling.
Most importantly, in addition to symbolic representations the model can also handle dense semantic representations, thus addressing one of the major shortcomings of previous scaling models.
We hypothesize that this will result in better results for text scaling, which we will investigate in the next section.

\section{Quantitative Evaluation}
\label{sec:eval}

We now present our new benchmarking data set for text scaling that extends the work of \citet{Proksch2010PositionSpeeches} by incorporating additional languages and data from another legislative term from the European Parliament (EP). 
Then we describe our evaluation setup and report the main results of our experiments, comparing the performance of Wordfish and SemScale when applied to scale the parties based on their members' speeches in the European Parliament.

\subsection{A new benchmarking set for political text scaling: European Parliament Speeches}

In our work, we follow the experimental design adopted by \citet{Proksch2010PositionSpeeches} when testing the Wordfish algorithm in different languages (English, French and German). As in their work, we collect speeches from the European Parliament website. We decided to extend the resource and the experimental setting used in this previous work, to test the validity of our findings across more languages (adding Italian and Spanish) and legislative terms (5th and 6th). To do so, we first crawled all individual speeches of all European Parliament representatives regarding the periods under study from the official website of the European Parliament\footnote{\url{http://www.europarl.europa.eu}} which cover 10 years of European politics (1999-2009). These are the only two legislative terms where the transcripts of the speeches are available online and the majority of them have been consistently translated.\footnote{For more details, see the European Parliament decision of 20 November 2012 on amendment of Rule 181 of Parliament's Rules of Procedure concerning verbatim reports of proceedings and Rule 182 concerning the audiovisual record of proceedings.} 

Unlike \citet{Proksch2010PositionSpeeches}, who considered all speeches from all MEPs in the English, French and German translations, we only keep speeches that have been \emph{originally} delivered in one of the five languages under study and translated to \emph{all} of the remaining four languages. I.e., we omit speeches delivered in some of the five languages that were not manually translated to each of the other four languages. This allows us to build {\em maximally comparable corpora} for all five languages, thus avoiding the issue of not always having a translation available for each language.\footnote{Note that this procedure of building maximally comparable corpora in all languages under study leads, however, to a data set which is different from the one used by \citet{Proksch2010PositionSpeeches} and the one used in our preliminary work \citep{Glavas2017UnsupervisedTexts}, where we consider all speeches available in any of the five languages.} 
Next, as done by Proksch and Slapin, we concatenate all speeches of all representatives of the same national party into a single party-level document for each language. Our data set (see statistics in Table \ref{tbl:data}), which we share together with this paper, represents a new resource for the evaluation of political text scaling algorithms in order to precisely examine their robustness across contexts and languages. However, it is also important to note that the difference in size between the two legislative corpora may have an impact on the results. %

\setlength{\tabcolsep}{10pt}
\begin{table}
\centering
\begin{tabular}{l c c c c}
\toprule 
Term & \# Parties & Min. Length & Mean Length & Max. Length\\ \midrule
5th (1999–2004) & 31  & 12K  & 160k & 543k \\
6th (2004–2009) & 26  & 11k  & 106k & 319k\\
\bottomrule
\end{tabular}
\caption{Statistics for the European Parliament data sets; number of words (computed on English subset of the data).
}
\label{tbl:data}
\end{table}

\subsection{Evaluation Setup for European Parliament Speeches}

Each unsupervised scaling technique assumes the existence of an underlying position/policy dimension across the documents under study. When processing transcripts of speeches from the European Parliament, \citet{Proksch2010PositionSpeeches} have shown that the dimension induced by Wordfish from EP speeches correlates better with parties' positions towards deeper EU integration than with their traditional Right-Left ideological positions. In this work, we replicate their analysis in order to validate their findings for Wordfish and test whether the same holds for semantically informed scaling with SemScale.

We follow \citet{Proksch2010PositionSpeeches} and consider as ground truth the positions of the parties under study derived from the Chapel Hill expert survey (years 2002 and 2006, respectively for the 5th and 6th legislative term)\footnote{\url{https://www.chesdata.eu/our-surveys/}} regarding the European integration process and a broad Right-Left ideology. 
We assess the quality of the scaling output by looking at the correlation of the results with  positions assigned by human experts. We compute the pairwise accuracy (PA), i.e., the percentage of pairs with parties in the same order, as well as Pearson  ($r_P$) and Spearman ($r_S$) correlation. While PA and Spearman correlation estimate the correctness of the ranking order, Pearson correlation also captures the extent to which the ground truth distances between party positions are correctly captured by the automatically induced scale. In the result tables, we report the average of each measure across the five languages under study. This will highlight how much the scaling results correlate with known positions of parties; a breakdown of the results for each language is available in the online appendix. Additionally, we present visual representations of the robustness of the inferred party positions across different languages. \\
 
\noindent \textbf{Parameter settings.}  In our experiments, we use the Quanteda implementation of Wordfish with default parameters.\footnote{For details, please refer to \url{https://quanteda.io/reference/textmodel_wordfish.html}.} 
When computing the document frequency matrix on the European Parliament data set, we empirically set a minimum document frequency of 5 on the basis of empirical tests and following findings and standard practices from previous work \cite{nanni2018,Bruinsma19,Egerod20}. For the experiments on the Manifestos in Section \ref{sub:sm}, we use a smaller threshold of $N=2$, motivated by the smaller corpus size of the data. The choice to empirically set a document frequency threshold for dimensionality reduction was motivated by \cite{yang97}, who showed that this approach has the following advantages: the technique scales well to large datasets and shows a high correlation with information gain and chi-square statistics.

For SemScale, different options can impact results:
(i) the type of preprocessing applied to the input documents (i.e., tokenization, lemmatization, filtering of input according to linguistic features, cf. Section \ref{sub:lf});
(ii) whether or not the input has been filtered by removing stopwords;
(iii) the type of embeddings used for computing the similarity between documents (please note that for optimal results, the preprocessing of the input documents should match the preprocessing applied to the corpus used for computing the word embeddings);
(iv) the type of similarity function used for computing document similarities (as described above, we use the cosine function);
(v) the graph-based clustering algorithm (we use the HFLP algorithm in all experiments and present a comparison with PageRank in Section \ref{sub:gca}).

As \citet{Denny2018TextIt} have recently highlighted, virtually any type of text preprocessing has a major impact on the scaling output (i.e., party positions) produced by Wordfish. For this reason, we have decided to first evaluate both Wordfish and SemScale on the original texts, with standard preprocessing (i.e., the removal of stopwords and punctuation), but applying neither stemming nor lemmatization to the input texts.
While this setting might not be optimal for either of the algorithms, it allows us to compare the capabilities of the two scaling methods in isolation, avoiding the risk of incorrectly attributing performance differences that are due to some text preprocessing step to either of the algorithms. Consequently, in all other validation experiments in Section \ref{sub:lf}, in which we retain only some subset of the original texts (e.g., only nouns or only named entities), we explicitly make sure that both scaling algorithms receive exactly the same textual input.

\subsection{Results on European Parliament Speeches}

\setlength{\tabcolsep}{13pt}
\begin{table}
\centering
\begin{tabular}{l ccc ccc }
\toprule
 & \multicolumn{3}{c}{5th Leg} & \multicolumn{3}{c}{6th Leg} \\
\cmidrule(lr){2-4}
\cmidrule(lr){5-7}

& PA & $r_P$ & $r_S$ & PA & $r_P$ & $r_S$ \\
\midrule

Wordfish & 0.54 & 0.15 & 0.12 & 0.53 & 0.16 & 0.09 \\
&  (0.01) & (0.04) & (0.03) & (0.03) & (0.06) & (0.09) \\ 
 
SemScale & \textbf{0.60} & \textbf{0.32} & \textbf{0.27} & \textbf{0.59} & \textbf{0.29} & \textbf{0.27} \\
& (0.02) & (0.03) &  (0.05) & (0.02) & (0.03) & (0.06) \\ 

\bottomrule
\end{tabular}
\caption{Correlation of automatically induced positions (averaged over all five languages), using the entire text, with the ground truth \textbf{positions on EU integration}. Standard errors are in brackets.}
\label{tbl:eu-int}
\end{table}

\setlength{\tabcolsep}{13pt}
\begin{table}
\centering
\begin{tabular}{l ccc ccc}
\toprule
& \multicolumn{3}{c}{5th Leg} & \multicolumn{3}{c}{6th Leg} \\
\cmidrule(lr){2-4}
\cmidrule(lr){5-7}

& PA & $r_P$ & $r_S$ & PA & $r_P$ & $r_S$ \\ \midrule
Wordfish  & \textbf{0.54} & 0.04 & \textbf{0.11} & 0.50 & 0.07 & -0.04 \\ 
& (0.0)  &  (0.01) & (0.01) & (0.01) & (0.03) & (0.03) \\
SemScale & 0.52 & \textbf{0.22} & 0.05 & \textbf{0.54} & \textbf{0.11} & \textbf{0.10} \\ 
& (0.04) & (0.04) & (0.12) & (0.01) & (0.05) & (0.04) \\
\bottomrule
\end{tabular}
\caption{Correlation of automatically induced positions (averaged over all five languages), using the entire text, with the ground truth \textbf{Right-Left positions}. Standard errors are in brackets.\newline}
\label{tbl:ideo}
\end{table}

\begin{figure}[t]
\vspace{-1.5cm}
\minipage{0.33\textwidth}
  \includegraphics[scale=0.33,bb=0 0 640 480]{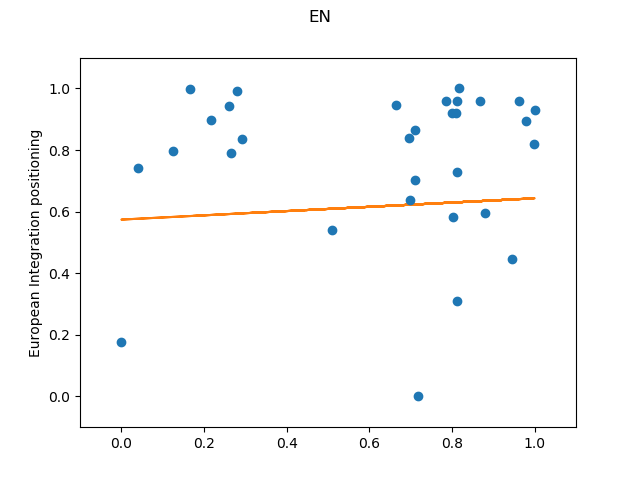}
\endminipage\hfill
\minipage{0.33\textwidth}
  \includegraphics[scale=0.33,bb=0 0 640 480]{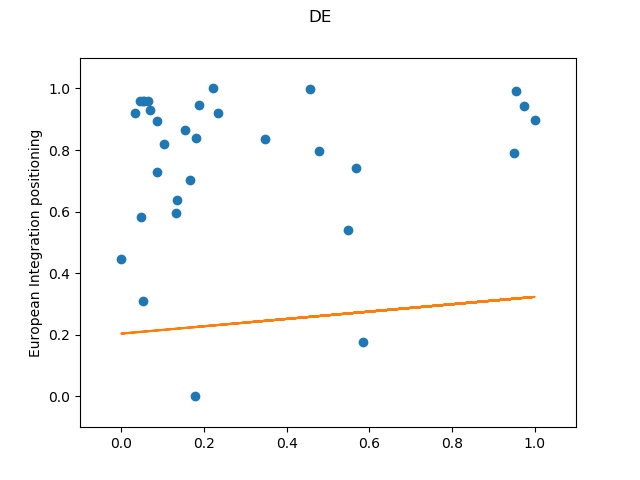}
\endminipage\hfill
\minipage{0.33\textwidth}
  \includegraphics[scale=0.33,bb=0 0 640 480]{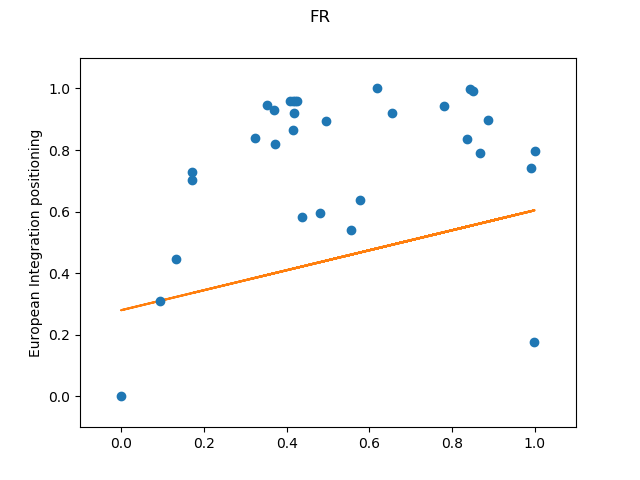}
  \endminipage
 \vspace{-1.5cm} 
\minipage{0.33\textwidth}
  \includegraphics[scale=0.33,bb=0 0 640 480]{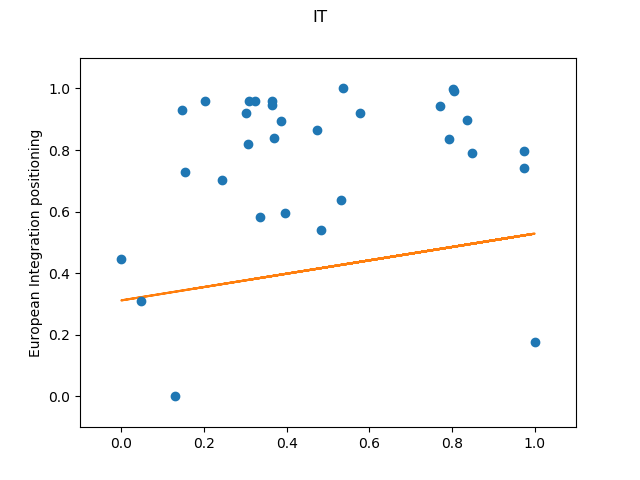}
\endminipage\hfill
\minipage{0.33\textwidth}
  \includegraphics[scale=0.33,bb=0 0 640 480]{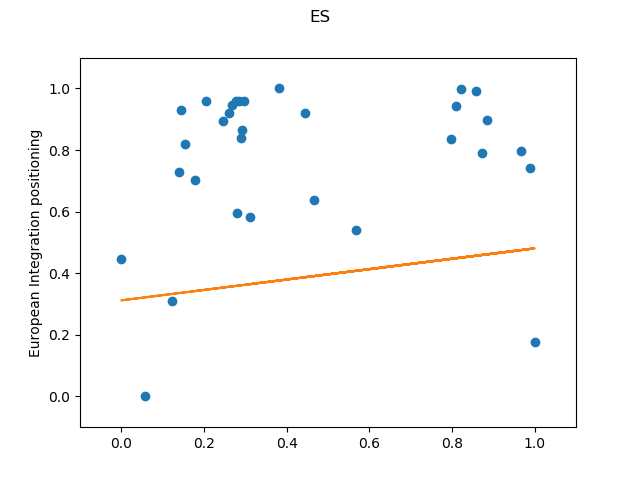}
\endminipage

\caption{Correlation of Wordfish results using the entire text (5th legislative term) with European Integration positioning.}
\label{fig:wordfish-5}
\end{figure}

\begin{figure}[t]
\vspace{-1.5cm}
\minipage{0.33\textwidth}
  \includegraphics[scale=0.33,bb=0 0 640 480]{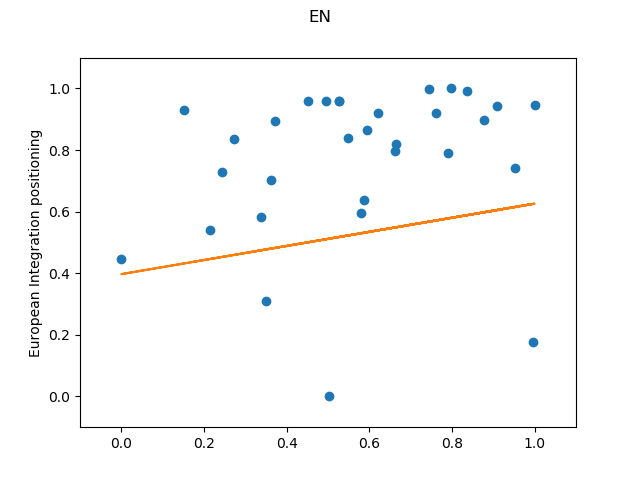}
\endminipage\hfill
\minipage{0.33\textwidth}
  \includegraphics[scale=0.33,bb=0 0 640 480]{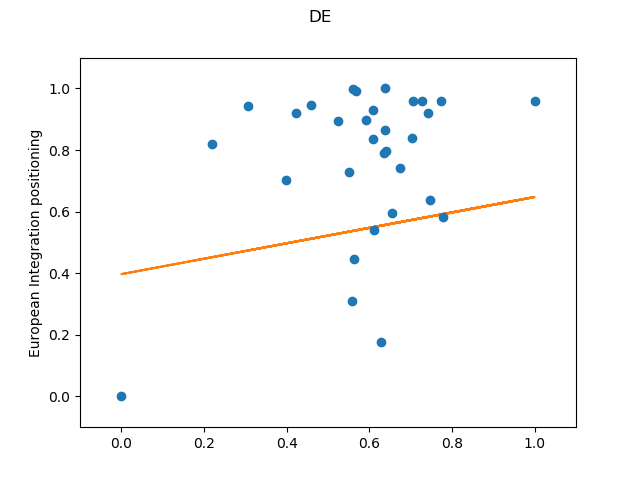}
\endminipage\hfill
\minipage{0.33\textwidth}
  \includegraphics[scale=0.33,bb=0 0 640 480]{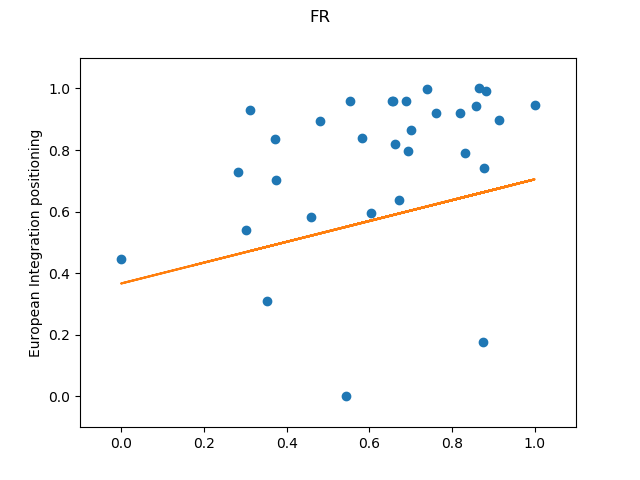}
  \endminipage 
 \vspace{-1.5cm} 
\minipage{0.33\textwidth}
  \includegraphics[scale=0.33,bb=0 0 640 480]{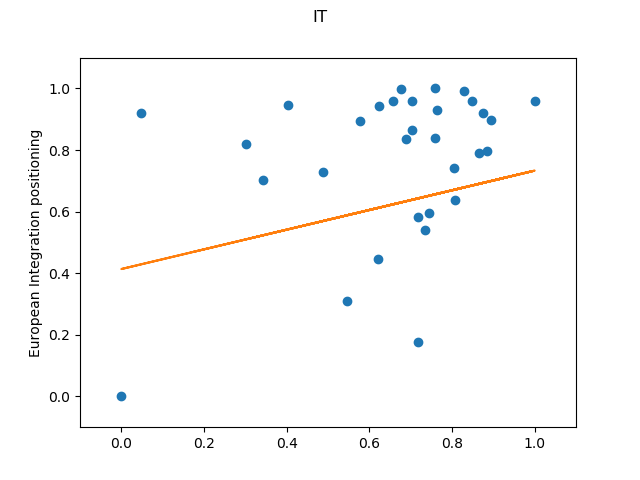}
\endminipage\hfill
\minipage{0.33\textwidth}
  \includegraphics[scale=0.33,bb=0 0 640 480]{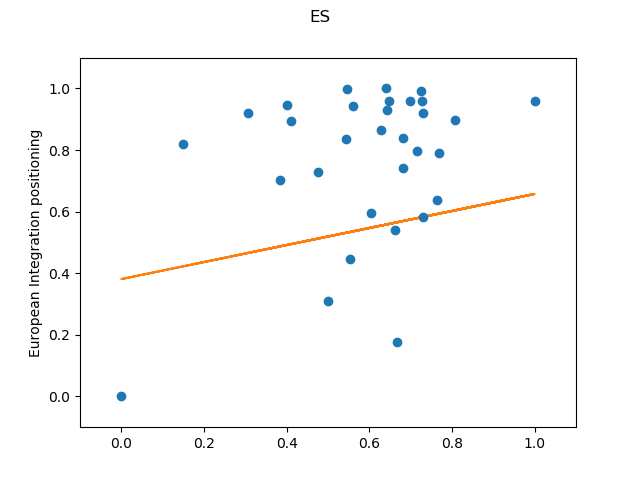}
\endminipage

\caption{Correlation of SemScale results using the entire text (5th legislative term) with European Integration positioning.}
\label{fig:sem-scale-5}

\end{figure}

In Table \ref{tbl:eu-int}, we present the averaged quality of the correlation between the produced scalings and the positions of the parties on the issue of European integration, according to the Chapel Hill Expert Survey, for the two legislative terms under study. 
The numbers clearly show that the positions induced by SemScale have a significantly higher correlation with ground truth (i.e., Chapel Hill) party positions on EU integration than the positions induced by Wordfish.   
The results, consistent across parliaments and languages, also confirm the findings of \citet{Proksch2010PositionSpeeches}, namely that scalings produced by Wordfish employing the entire text correlate better with the parties' positions concerning European integration than the ideological Right-Left dimension (cf.~Table \ref{tbl:ideo} for comparison). Moreover, they highlight that this effect is even more prominent when adopting SemScale, a semantics-aware text scaling algorithm. These findings are further emphasised by Figures \ref{fig:wordfish-5} and \ref{fig:sem-scale-5}, which reveal the high level of consistency of SemScale across languages. 
This emphasises the complexity of evaluating the output of text scaling algorithms and the importance of always considering more than a single (ideological) dimension when interpreting their output. 

In the next section, we expand on this and present a number of validation experiments where we explore the robustness of the scaling algorithms and systematically evaluate the impact of the different components of our model, i.e., the text representation module and the graph-based clustering algorithm.

\section{Validation experiments}
\label{sec:validation}

In the last section, we have shown that SemScale outperforms Wordfish on scaling party positions on European integration across five languages and two legislative terms. 
To further validate the effectiveness of SemScale and to better understand its potential and limitations, we now present a series of experiments concerning the impact of text preprocessing (\ref{sub:lf}, {\sc Preprocessing: The Impact of Linguistic Features}) and text representations (\ref{sub:dtr}, {\sc Different Text Representations: tf-idf and Party2Vec}) on scaling performance. 
We test the robustness of SemScale by applying it to a different type of political text (\ref{sub:sm}, {\sc Scaling Manifestos}) and assess the influence of different graph clustering algorithms on scaling results (\ref{sub:gca}, {\sc Graph Clustering Algorithms: HFLP versus PageRank}).
In our final validation experiments, we extend Wordfish with word embeddings to test whether this can further improve results for text scaling  (\ref{sub:wwwe}, {\sc Wordfish with Word Embeddings}) and adapt our scaling algorithm to a supervised setup in which ground truth positions are available for some of the texts (\ref{sub:wss}, {\sc Weakly Supervised Scaling}).

\subsection{Preprocessing: The Impact of Linguistic Features}
\label{sub:lf}

In previous sections, we have reported on criticism raised by \citet{Denny2018TextIt} who investigated the impact of preprocessing on text scaling and showed that the widely-used scaling algorithm Wordfish \citep{slapin_scaling_2008} is not very robust: even small and semantically insignificant text preprocessing steps may have a profound impact on the scaling results. 
Extending their work, we examine the impact of parts of speech (POS) tagging and named entity recognition on the obtained scaling and its correlation with EU integration positioning. 
In other words, we quantify how stable different scaling algorithms are with respect to different preprocessing and content selection steps. 

\begin{figure}
 \begin{center}
  \includegraphics[width=\textwidth]{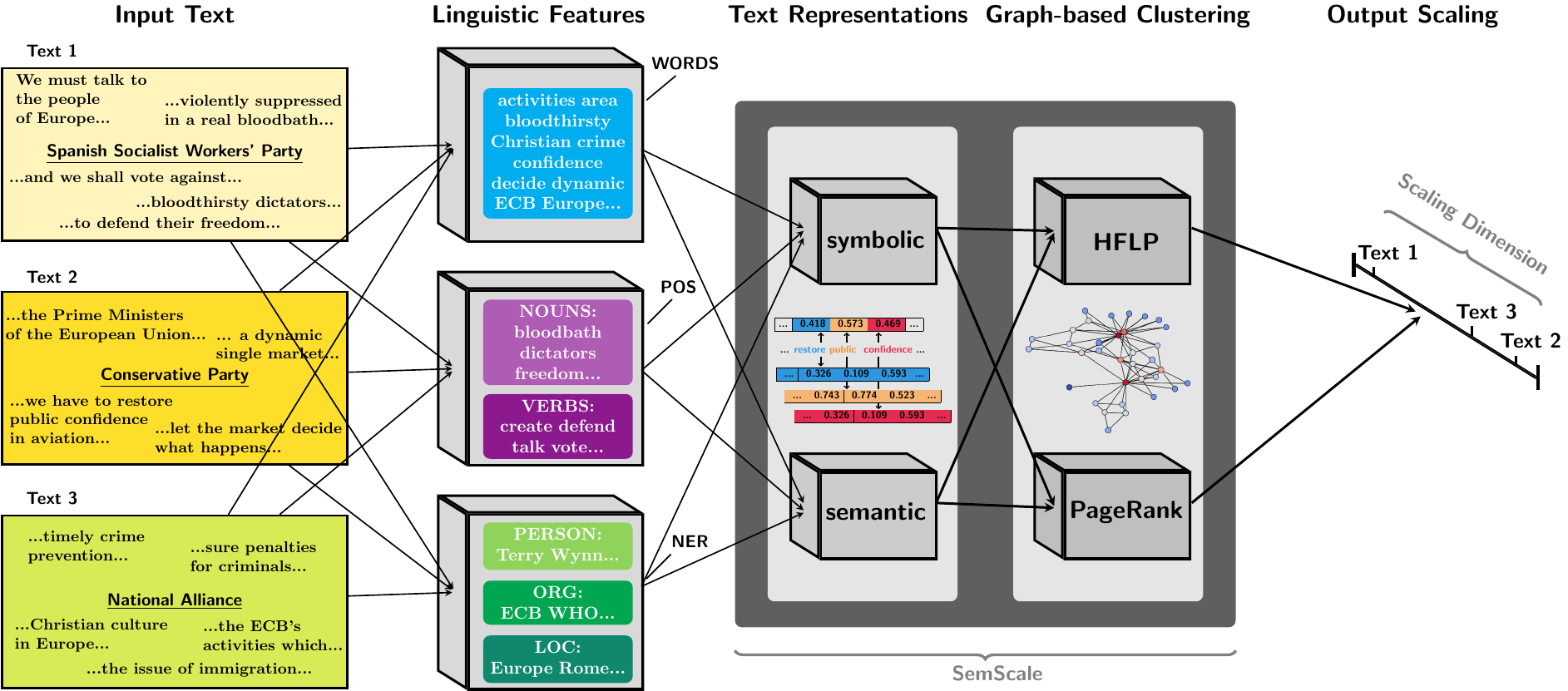}
 \end{center}
\caption{Illustration of the use of different linguistic features in SemScale and options for obtaining the final scaling, based on combinations of different text representations and clustering algorithms (e.g., nouns only + semantic text representations + HFLP; {\em or} verbs and person names + symbolic text representations + PageRank).}\label{fig:feats}
\end{figure}

We are particularly interested in understanding whether the positions produced by scaling algorithms show a higher correlation with specific subsets of linguistic traits and whether this could point to text preprocessing steps which facilitate scaling and lead to better predictions of party positions.
To do so, we filter the scaling input and only keep words of a specific part of speech or named entity type, and then scale the input documents, based on the selected linguistic features (see Figure \ref{fig:feats}).\\

\noindent \textbf{Parts of Speech.} The computational linguistics community has put a lot of effort into developing systems capable of identifying the different parts of speech for words in text (nouns, verbs, adjectives, etc.). While older generations of POS tagging models have been based on traditional sequence tagging algorithms such as Hidden Markov Models (which assume that a sequence of words is generated by a Markov process where the parts of speech are latent states that need to be uncovered) or Conditional Random Fields \citep{Lafferty2001ConditionalData}, state-of-the-art POS tagging models are commonly built upon deep neural networks.\footnote{For instance, bidirectional recurrent neural networks or residual convolutional networks, cf. \citet{goodfellow2016deep} for a comprehensive overview of deep learning architectures.} 
In order to test the effects that different parts of speech have on the scaling procedure, we have used a POS tagger to filter for nouns and verbs in the input texts, for each of the five languages under consideration. \\

\noindent \textbf{Lemmas.} A common practice in computational text analysis is the normalisation of the texts under study, to reduce the overall vocabulary via \textit{morphological normalisation}. During this process, all different morpho-syntactic forms of the same word (e.g., \textit{``house''}, \textit{``houses''}, \textit{``housing''}) are reduced to some common form (e.g., \textit{\textit{``house''} or \textit{``hous''}}). The most common types of morphological normalization are (1) stemming \citep[e.g.][]{Porter1980AnStripping}, 
which strips the suffixes from the word, based on a series of heuristics and predefined rules, and (2) lemmatization, which reduces different word forms to their canonical form (e.g., cases of nouns to singular nominative or different conjugations of a verb to its infinitive form). 

Stemming has already been shown to have a negative impact on automatic text scaling \citep{Greene2016TheCountries,Denny2018TextIt} as it may add lexical ambiguity (e.g., \textit{``party''} may be stemmed to \textit{``part''}, which holds a different meaning). 
Lemmatization, instead, normalises inflected word forms to their canonical forms called lemmas (e.g., \textit{``parties''} to \textit{``party''} or \textit{``voted''} to \textit{``vote''}). Lemmatization is often performed by a look-up in a dictionary of inflected forms, using the inflected word and its part of speech as look-up keys (e.g., the POS information helps to transform \textit{``meeting''} to \textit{``meet''} when used as a verb and leave it unchanged when used as a noun). 
When using lemmas instead of the words themselves, we obtain lemma embeddings by first (1) lemmatizing all European Parliament speeches (i.e., we create the lemmatised in-domain text corpus) and then (2) running a word embedding algorithm on the previously lemmatised corpus. 

In our experiments, lemmatization does not seem to have a significant effect on scaling accuracy for both, WordFish and SemScale. We obtained similar correlations with expert positions when using lemmatised and non-lemmatised texts. Thus, in the paper we only report results for non-lemmatised texts.
However, a major advantage of lemmatization is that it helps to reduce the overall size of the word vocabulary and consequently speeds up the automatic scaling process. 
\\ 

\noindent \textbf{Named Entities.} When dealing with large amounts of text documents, a useful strategy for finding relevant pieces of information is to highlight all named entities that are mentioned in the text. Similar to POS tagging, named entity recognition (NER) is a sequence labelling task where, for each word in the text, a label is created that signals whether this word is part of a named entity (e.g., of type PERSON, ORGANISATION, or LOCATION). 
As for POS tagging, previous generations of named entity recognition systems were also based on machine learning models such as the Hidden Markov Model and Conditional Random Fields, whereas the most recent NER approaches use deep neural networks \citep[see, for instance,][]{ganea2017deep} or hybrid systems that combine neural network with Conditional Random Fields \citep{lample:etal:2016,ma:hovy:2016}. In any case, a large corpus of text manually annotated with named entities is required to train a reliable NER model. \\

\noindent \textbf{Tools Adopted.} One of the main goals of our empirical methodology was to use computational approaches that are as comparable as possible across different languages; for this reason, whenever possible, we adopted the same infrastructure, models, and tools for linguistic analysis for each of the five languages. For POS tagging, lemmatization and named entity recognition we employed spaCy,\footnote{\url{https://spacy.io/}} a Python library that offers robust pre-trained models for all five languages under study.\footnote{We initially considered using Stanford CoreNLP by \citet{Manning2014TheToolkit}, a more widely adopted natural language toolkit. However, we found models for all required tasks -- POS-tagging, lemmatization, and NER -- only for English, Spanish, and German.}  
We computed word embeddings on the European parliament speeches and political manifestos, using the FastText word embedding tool \citep{Bojanowski2017EnrichingInformation}.\footnote{We used the Gensim implementation of FastText (\url{https://radimrehurek.com/gensim_3.8.3/}) with the skipgram algorithm, a dimension size of 300, a window size of five and a minimum token count of five for the in-domain embeddings.} 
We make all the resources and preprocessed data sets available for further research in our online supplementary materials.

Before we report our results, we would like to remark that while the above-mentioned tools are widely adopted by both the academic and industrial computational linguistics communities, their performance, especially, for more complex tasks (Named Entity Recognition) is far from optimal.\footnote{As also documented by spaCy itself: \url{https://spacy.io/usage/facts-figures}} Nevertheless, with the aim of opening the discussion and motivating further research efforts on using semantic enrichment of text for political text scaling, we have employed these models with the awareness of their current limitations. 
By demonstrating that even with their current, sub-ideal performance these models can significantly contribute to the scaling quality, we believe that with future advances in computational linguistics, we will also witness further improvements in political text scaling. \\

\begin{table}
\centering
\begin{tabular}{ll ccc ccc }
\toprule
& & \multicolumn{3}{c}{5th Leg} & \multicolumn{3}{c}{6th Leg} \\
\cmidrule(lr){3-5}
\cmidrule(lr){6-8}
& & PA & $r_P$ & $r_S$ & PA & $r_P$ & $r_S$ \\
\midrule
 Wordfish & {\sc noun} & 0.54 & 0.14 & 0.11 & 0.57 & 0.24 &	0.18 \\
SemScale & {\sc noun} & \textbf{0.55} & \textbf{0.23} & \textbf{0.14} & \textbf{0.58} & \textbf{0.25} & \textbf{0.22} \\
\midrule
 Wordfish & {\sc verb} & 0.55 & 0.18 & 0.15 & \textbf{0.58} & \textbf{0.28} & \textbf{0.25} \\
SemScale & {\sc verb} & \textbf{0.65} & \textbf{0.36} & \textbf{0.46} & 0.55 & 0.22 & 0.15 \\
\bottomrule

\end{tabular}
\caption{Correlation of automatically induced positions with the ground truth {\bf positions on EU integration} when we use \textbf{only nouns/verbs} from the original text as input for text scaling.\newline}
\label{tbl:eu-int-nouns-verbs}
\end{table}

\begin{table}[t]
\centering
\begin{tabular}{ll ccc ccc }
\toprule
& & \multicolumn{3}{c}{5th Leg} & \multicolumn{3}{c}{6th Leg} \\
\cmidrule(lr){3-5}
\cmidrule(lr){6-8}
& & PA & $r_P$ & $r_S$ & PA & $r_P$ & $r_S$ \\
\midrule
Wordfish & {\sc person} & 0.55 & 0.11 & 0.12 & 0.63 & 0.42 & 0.37 \\
SemScale & {\sc person} & {\bf 0.62} & {\bf 0.38} & {\bf 0.31} & {\bf 0.66} & {\bf 0.49} & {\bf 0.46} \\

Wordfish & {\sc org}  & {\bf 0.58} & {\bf 0.28} & {\bf 0.22} & 0.60 & {\bf 0.35} & 0.29 \\

SemScale & {\sc org} & {\bf 0.58} & 0.18 & 0.21 & {\bf 0.61} & 0.31 & {\bf 0.31} \\
\bottomrule
\end{tabular}
\caption{Correlation of automatically induced positions with {\bf European Integration positioning} when \textbf{only person names} or \textbf{only organisations} are used for text scaling.\newline}  
\label{tbl:eu-int-per-org}
\end{table}
 
\noindent\textbf{Nouns and verbs}. In Table \ref{tbl:eu-int-nouns-verbs}, we show how positions produced by Wordfish and SemScale correlate with ground truth positions on EU integration when we use only the nouns or only the verbs from the input texts, respectively. 
For SemScale, noun-only results decrease as compared to scaling on the entire text, but are still higher than the ones for Wordfish. Scaling on verbs only increases results for the 5th legislative term but not for the 6th for SemScale, while for Wordfish the results are in the same range or higher. \\

\noindent\textbf{Named Entities}. We next move to the analysis of scaling results produced when the input to the scaling algorithms consists of names only (i.e., mentions of named entities such as persons, organisations, locations etc.).  
When scaling documents based only on the {\em person names} in the texts, SemScale produces positions that show a high correlation  with party positions on EU integration (Table \ref{tbl:eu-int-per-org}). 
In contrast, scaling based on mentions of {\em organisations} (e.g., \textit{Euratom}, \textit{Parmalat}, \textit{PKK}) produces results that are less consistent (see Table \ref{tbl:eu-int-per-org}). 
We believe that this is primarily due to the variance in performance for named entity recognition models for different languages and named entity types: Typically, person names are easier to recognise across languages while organisations are much harder to disambiguate, thus leading to inconsistent results. This, of course, may result in semantically very different inputs to the scaling algorithms for different languages.\\

When evaluating the predicted positions against the Right-Left ideological dimension (not shown here), our results again confirm the original findings from \citet{Proksch2010PositionSpeeches} that the EU integration dimension is clearly more prominently captured by text scaling methods than the Right-Left ideological dimension.

Overall, our experiments confirm the findings of \citet{Denny2016AssessingDecisions} and show that both Wordfish and SemScale are sensitive to changes in the input data.
Our results also show that informed preprocessing, i.e., filtering the input based on linguistic traits, can improve results over using the whole document content as input (cf. Table \ref{tbl:eu-int}). However, we need to better understand when preprocessing is beneficial and when it might harm the results.
We hope our findings and the data sets we release will motivate further research on the role that entities such as person names, organisations or locations play in deducing ideological positions from textual data.

\subsection{Different Text Representations: TF-IDF and Party2Vec}
\label{sub:dtr}

In the last section, we looked at different ways to filter the input for text scaling, based on linguistic preprocessing. Here, we investigate the impact of different input representations on scaling results: {\em a)} symbolic representations (i.e., word counts), {\em b)} semantics-aware representations (averaged word embeddings) and {\em c)} distributional representations of documents (Party2Vec).

Our SemScale approach consists of two main components: (1) a function that measures the similarity between individual documents as the similarity between {\sc tf-idf}-weighted averages of word embeddings (from now we will refer to this similarity function as \textsc{Avg w-emb}) and (2) a graph-based scaling algorithm that operates on the similarity graph as input, regardless of how the similarity scores were computed.  
This allows us to easily plug in and test different input representations, as the ones named above.

First, instead of using the \textsc{Avg w-emb} as in previous experiments, we adapt SemScale to operate with a simple symbolic (i.e., sparse) document representation: a term frequency–inverse document frequency (\textsc{tf-idf}) vector for each document. Such representation, which captures word frequency information and does not model semantics, is widely used in traditional text mining as a baseline method in many tasks, from text classification and clustering to information retrieval and regression analyses \citep{manning_foundations_1999}. In this case, the similarity function for the construction of the similarity graph is simply the cosine similarity between the sparse \textsc{tf-idf} vectors of documents.  
Secondly, we investigate a different text representation employing the recently proposed party embeddings \citep{rheault2019word}. In this work, an algorithm for directly inducing document embeddings, dubbed Doc2Vec \citep{le2014distributed}, is used to obtain vector representations of parties: the whole procedure is here referred to as \textsc{Party2Vec}. We have built party embeddings following the procedure and code\footnote{\url{https://github.com/lrheault/partyembed}} from \citet{rheault2019word} and used the cosine similarity between the obtained party vectors as the similarity function with which we induce the similarity graph for SemScale's graph-based scaling algorithm.

\begin{table}
\centering
\begin{tabular}{l ccc ccc}
\toprule
 & \multicolumn{3}{c}{5th Leg} & \multicolumn{3}{c}{6th Leg} \\
\cmidrule(lr){2-4}
\cmidrule(lr){5-7}

& PA & $r_P$ & $r_S$ & PA & $r_P$ & $r_S$ \\ \midrule

Wordfish  & 0.54 & 0.15 & 0.12 & 0.53 & 0.16 & 0.09 \\
&  (0.01) & (0.04) & (0.03) & (0.03) & (0.06) & (0.08) \\
\midrule
\textsc{tf-idf} &  0.58 & 0.22 & 0.23 & 0.57 & 0.20 & 0.19 \\
& (0.03) & (0.07) & (0.07) & (0.03) & (0.08) & (0.1) \\

\textsc{Avg w-emb} & \textbf{0.60} & \textbf{0.32} & \textbf{0.27} & 0.59 & 0.29 & 0.27 \\
 & (0.02) & (0.03) & (0.05) & (0.02) & (0.03) & (0.06)\\ 

 \textsc{Party2Vec} & 0.55 & 0.12 & 0.13 & \textbf{0.64} & \textbf{0.39} & \textbf{0.38} \\ 
&  (0.02) & (0.03) & (0.05) & (0.03) & (0.09) & (0.08)\\   
\bottomrule
\end{tabular}
\caption{Correlations of automatically induced positions with ground truth positions on EU-Integration across different languages. Comparison between using \textsc{tf-idf, Party2Vec}, and averaged word-embeddings as different text representations for SemScale's graph-based scaling algorithm. Standard errors are in brackets. Wordfish performance is reported as a point of reference.}
\label{tbl:vectors}
\end{table}

The results shown in Table \ref{tbl:vectors} denote the correlation scores with European Integration positioning (averaged across five languages) when using different text representations and, consequently, different similarity functions as input for SemScale's graph-based scaling. 
It is important to notice that SemScale's performance when relying on sparse symbolic representation of text (i.e., sparse \textsc{tf-idf} vectors) is still above the one of Wordfish but significantly below SemScale's performance when relying on word embeddings (\textsc{Avg w-emb}). 
For the 5th legislative term, we do not observe any improvement in results when replacing the document representations obtained by averaging word embeddings (\textsc{Avg-w-emb}) with \textsc{Party2Vec} embeddings \citep{rheault2019word}. On the contrary, the aggregation of word embeddings seems to provide a better semantic signal for scaling than the \textsc{Party2Vec} embeddings, but also shows a high variance in results, as compared to \textsc{Avg-w-emb} and \textsc{tf-idf}.  
For the 6th term, however, the \textsc{Party2Vec} embeddings seem to provide a stronger and more reliable signal for text scaling.   

Our results show that the use of word embeddings is a core component of SemScale and should not be substituted by (symbolic) word frequency information alone. 
However, the results also show that the success of SemScale can not be explained by the use of semantics-aware word embeddings alone but that the graph-based scaling approach might also play a role, given that results for SemScale with {\sc tf-idf} outperform Wordfish on both data sets from the 5th and 6th legislative term. 
In addition, we confirm again that -- as noted by \citet{Denny2018TextIt} and in our own experiments in the last section -- text scaling algorithms can be highly sensitive to small changes in the input signal.

\subsection{Scaling Manifestos}
\label{sub:sm}

As a further evaluation of the potential of SemScale, we now test it on a different source of political texts, namely party manifestos from the Comparative Manifestos Project\citep{Merz2016TheAnalysis}.\footnote{The data is available from \url{https://manifestoproject.wzb.eu/}.} We collect all electoral manifestos from the United Kingdom, Germany, France, Italy and Spain which are available with manifesto coded annotations at the quasi-sentence level (i.e., a sentence or clause). We do this because, while we do not employ the annotations in our work, we aim to establish a common benchmark that future studies could employ to extend our work even if they intend to rely on the provided annotations.

\setlength{\tabcolsep}{10pt}
\begin{table}[t]
\centering
\begin{tabular}{l ccc ccc}
\toprule
 & \multicolumn{3}{c}{Single Election} & \multicolumn{3}{c}{Multiple Elections} \\
\cmidrule(lr){2-4}
\cmidrule(lr){5-7}

& PA & $r_P$ & $r_S$ & PA & $r_P$ & $r_S$ \\ \midrule

Wordfish & 0.63 &	0.44 &	0.35 & \textbf{0.61} & 0.26 & \textbf{0.31} \\ 
& (0.06) & (0.12) & (0.15) &  (0.06) & (0.09) & (0.16)  \\  

SemScale ({\sc Avg w-emb}) & 0.58 & 0.21 & 0.24 & 0.56 & 0.18  & 0.18 \\
&  (0.04) & (0.11) & (0.09) & (0.04) & (0.09) & (0.12) \\  

SemScale ({\sc Party2Vec}) & \textbf{0.65} & \textbf{0.46} & \textbf{0.38} & 0.59 & \textbf{0.30} & 0.27 \\
&  (0.06) & (0.12) & (0.14) & (0.02) & (0.06) & (0.06) \\ 
\bottomrule
\end{tabular}
\caption{Results for automatic text scaling of party manifestos: correlations of predicted positions with RILE (Right-Left ideology) ground truth scores from the Manifesto Project, across different countries. Comparison between positioning of manifestos from a single election and manifestos from multiple elections. Standard errors are in brackets.\newline}
\label{tbl:manifestosbycountry}
\end{table}

The manifestos are further divided into two data sets that we call "Single Election" and "Multiple Elections". The first includes electoral manifestos from only a single (recent) election for each country: UK 2015, Germany 2013, France 2012, Italy 2013, Spain 2011; the second data set contains all available coded manifestos for each country. For each country under study and for each data set, we measure the correlation between the produced scaling and the Right-Left ideological positioning (RILE) of the document provided by the Manifesto Project. 

Before we present the results, it is important to remark on a few aspects of this specific experiment: \emph{a)} in contrast to positions of European parties from the Chapel Hill Expert Survey (used as ground truth in previous experiments), which were not (at least not directly) assigned by expert annotators based on the content of the EP speeches, the RILE positions from the Manifesto project are derived directly from the coded quasi-sentences, meaning that they reflect the positions expressed by the texts themselves; \emph{b)} the Text as Data community has already discussed in detail many of the critical aspects and limitations of the Manifesto Project coding scheme and in particular of RILE.\footnote{See, \textit{inter alia}, \citep{mikhaylov2008coder,lacewell2013coder,gemenis2013and}.} While aware of the criticism, in this study we employ RILE scores because they are ground truth scores derived from the same text available to the automatic scaling methods, as opposed to the Chapel Hill positions, which are assigned by the experts based on their general familiarity and knowledge of political parties; 
\emph{c)} in our previous experiments with EP speeches, the texts in different languages were direct translations of each other. This, however, is not the case for the manifestos, meaning that the obtained results are therefore not directly comparable across countries / languages.

\begin{figure}[t]
        \center{\includegraphics[width=0.80\textwidth]
        {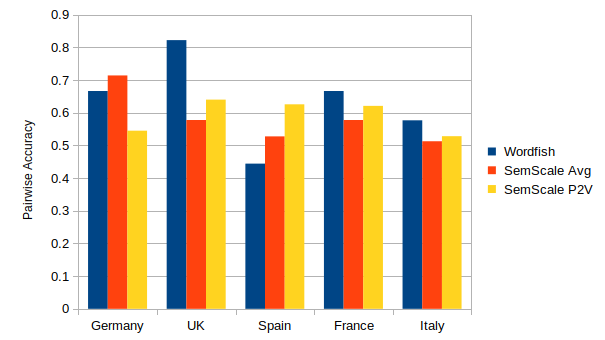}}
        \caption{\label{fig:manifestosbycountry} Pairwise accuracy of party positions produced on party Manifestos by different scaling algorithms and the RILE Right-Left ground truth ideological positions, across different countries.}
      \end{figure}

Scaling results on the Manifestos data set are displayed in Table \ref{tbl:manifestosbycountry}.\footnote{In the Manifestos experiments, we decided to decrease the threshold for minimum document frequency from 5 to 2 for Wordfish when computing the document frequency matrix, due to the smaller size of the dataset as compared to the data from the European Parliament. Accordingly, we use a minimum token count of 2 when computing the word embeddings for SemScale. Other parameter settings remain the same as in previous experiments.} 
First of all, we notice that Wordfish better predicts manifesto RILE positions, as compared to the default SemScale variant (with \textsc{Avg w-emb}).
SemScale, based on {\sc Party2Vec} embeddings, however, outperforms Wordfish in the single election setting, while results for the multiple election setting are mixed.
Figure \ref{fig:manifestosbycountry} offers a more detailed per-country view of the scaling results (in terms of the \textit{pairwise accuracy} measure). 
We can see that there is no single scaling method that yields best predictions in all five settings, i.e.,  for all countries. The performance of each method greatly varies, especially for Wordfish, and ranges from compelling 0.82 (when applied to UK manifestos) to mere 0.44 (on Spanish manifestos)\footnote{The PA performance of 0.44 is below a random predictions baseline of party positions which, on average, is expected to yield a PA of 0.5.} while SemScale (both based on \textsc{Avg-w-emb} and Party2Vec) tends to have more consistent correlations across countries and never performs below the random prediction baseline.

We speculate that the lower results for SemScore on the Multiple Elections data set is due to the large semantic shift in the content of such texts, spanning across decades of national politics, to which our semantic algorithm seems to be more sensitive than Wordfish. For these reasons, it is arguably not recommendable to use SemScale outside of the same temporal context (e.g., the same election, campaign, political debate, legislative period) as temporal shifts may severely affect the results: In such settings SemScale may find topical rather than positional similarities between documents and may assign similar positions to texts because they cover the same range of topics.

\subsection{Graph Clustering Algorithms: HFLP versus PageRank}
\label{sub:gca}

In the previous set of validation experiments, we have investigated changes concerning the input representations and content used for text scaling. We now present experiments designed to assess the importance of the choice of graph clustering algorithm used in SemScale. 
For that, we compare the performance of the \textit{harmonic function label propagation} (HFLP) algorithm that we used in our previous experiments to results obtained when using PageRank instead of HFLP.

Originally developed by Google for ranking search results in web searches, PageRank counts the number of links to a retrieved document to estimate its importance. This is done with the help of {\em random walks} that follow the links between the websites, where the rank of each document is based on the probability of landing on that particular website, which will be higher for important sites with many incoming links. This approach can also be interpreted as a Markov process where each step of the random walk only depends on the current state, but not on the history (i.e., the previous steps).

Besides web searches, the PageRank algorithm can also be applied to many other problems in NLP (see, e.g., \cite{agirre:soroa:2009,pershina:etal:2015,wachsmuth:etal:2017}). In order to use PageRank for text scaling, we create our weighted similarity graph as described before, where each document is a node in the graph and the similarity between documents are weighted edges between the nodes.
Thus, we can consider the similarity between the documents, based on averaged word embeddings, as ``links'' from one document to another, with each document (or node) representing the texts produced by one particular party. The likelihood of taking a random walk and landing on a document that is similar to the one from which we started is thus much higher than landing on a document that is completely different to our starting point.

Accordingly, in the next experiment, we use PageRank to compute a probability distribution that represents the likelihood of randomly landing on a particular document when taking the next step in a random walk. When starting the walk, the probability distribution is divided evenly among all documents, and in each iteration of learning the values are adjusted, getting more and more accurate while the learning proceeds.  

\setlength{\tabcolsep}{13pt}
\begin{table}[t]
\centering
\begin{tabular}{l ccc ccc }
\toprule
 & \multicolumn{3}{c}{5th Leg} & \multicolumn{3}{c}{6th Leg} \\
\cmidrule(lr){2-4}
\cmidrule(lr){5-7}

& PA & $r_P$ & $r_S$ & PA & $r_P$ & $r_S$ \\
\midrule 
EP SemScale HFLP & \textbf{0.60} & \textbf{0.32} & \textbf{0.27} & 0.59 & {0.29} & {0.27} \\
& (0.02) & (0.03) &  (0.05) & (0.02) & (0.03) & (0.06) \\ 
EP SemScale PageRank & 0.54 & 0.28 & 0.09 & \textbf{0.67} & \textbf{0.43} & \textbf{0.47} \\
& (0.01) & (0.03) & (0.02) & (0.01) & (0.03) & (0.03) \\
\bottomrule
\end{tabular}
\caption{Correlation of automatically induced positions (averaged over all five languages),  with the ground truth \textbf{positions on EU integration}.}
\label{tbl:hflp-pr-eu} 
\end{table}

\setlength{\tabcolsep}{13pt}
\begin{table}[t]
\centering
\begin{tabular}{l ccc ccc }
\toprule
 & \multicolumn{3}{c}{Single Election} & \multicolumn{3}{c}{Multiple Elections} \\
\cmidrule(lr){2-4}
\cmidrule(lr){5-7}

& PA & $r_P$ & $r_S$ & PA & $r_P$ & $r_S$ \\ \midrule
Manifestos SemScale HFLP & 0.58 & 0.21 & 0.24 & \textbf{0.56} & \textbf{0.18}  & \textbf{0.18} \\
&  (0.04) & (0.11) & (0.09) & (0.04) & (0.09) & (0.12) \\

Manifestos SemScale PR & \textbf{0.60} & \textbf{0.35} & \textbf{0.30} & 0.50 & 0.10 & 0.01 \\
& (0.04) & (0.10) & (0.12) & (0.03) & (0.05) & (0.09) \\
\bottomrule
\end{tabular}
\caption{Correlation of automatically induced positions (averaged over all five languages) with RILE (Right-Left ideology) ground truth scores from the Manifesto Project. Standard errors are in brackets.}
\label{tbl:hflp-pr-manifestos} 
\end{table}

Table \ref{tbl:hflp-pr-eu} shows results for the two graph-based clustering algorithms on the documents from the EU parliament for correlations with positions on EU integration, and table \ref{tbl:hflp-pr-manifestos} presents the same results for predicted party positions, evaluated against ground-truth positions from RILE on party manifestos. We can see that the choice of algorithm has a huge impact on results, and  that no algorithm provides best results for all settings. However, it seems as if HFLP provides a more reliable and robust signal, as shown by consistent Spearman ($r_S$) correlation scores, while PageRank only shows a very weak correlation or none at all in two of the settings ($r_S$<0.1 for EuroParl 5th leg. and $r_S$=0.01 for Manifestos, Multiple Elections).
We thus recommend using SemScale with HFLP instead of PageRank.

 \begin{table}
\centering
\begin{tabular}{l ccc r ccc r}
\toprule
 & \multicolumn{3}{c}{5th Leg} & voc. & \multicolumn{3}{c}{6th Leg} &  voc. \\
\cmidrule(lr){2-5}
\cmidrule(lr){6-9}
& PA & $r_P$ & $r_S$ & & PA & $r_P$ & $r_S$ & \\ \midrule 
 Wordfish & 0.54 & 0.15 & 0.12  & 66k   & 0.53 & 0.16 & 0.09  & 53k \\
&  (0.01) & (0.04) & (0.03)     &    & (0.03) & (0.06) & (0.08) & \\
\midrule
Wordfish+E  & 0.55 & 0.18 & 0.12 & 33k	&  0.54 &	0.20 & 0.11 & 24k\\
($\theta$: 0.95) & (0.01) & (0.06) & (0.03) &    & (0.03) & (0.06) & (0.9) & \\
\midrule
Wordfish+E & 0.55 & 0.16 & 0.13 & 11k    & 0.54 &	0.21 & 0.12 & 7k \\
 ($\theta$: 0.85) & (0.02) & (0.04) & (0.03) &    & (0.02) & (0.05) & (0.7) & \\
\midrule
Wordfish+E & 0.55	& 0.15 & 0.13 & 5k    & 0.54 &	0.19 &	0.11 & 3k \\
 ($\theta$: 0.80) & (0.01) & (0.03) & (0.03) &    & (0.03) & (0.06) & (0.8) & \\
\bottomrule
\end{tabular}
\caption{Correlations of automatically induced positions with ground truth positions on EU-Integration across different languages. Comparison between using original Wordfish and a version of Wordfish where we compute the document frequencies over groups of similar words, grouped based on the cosine similarity between the word embeddings (Wordfish+E). The threshold ($\theta$) specifies the minimum cosine similarity score needed for two words to be grouped together and voc.\ reports the avg.\ vocabulary size over all languages. Standard errors are in brackets.}\label{tab:wordfish-embed}
\end{table}
\subsection{Wordfish with Word Embeddings}
\label{sub:wwwe}

Given the crucial role of word embeddings for text scaling, we next investigate whether we can improve results for Wordfish by adding information on word similarity to the input. 
As already mentioned above, the main disadvantage of symbolic word representations is their sparsity and the fact that similar words are treated the same way as 
words with completely different meanings.

To adapt the Wordfish input so that similar words now have similar representations, 
we preprocess the input documents and, instead of counting word frequencies for individual word forms, we group words into semantic classes, based on the cosine similarity of their embeddings. We test different similarity thresholds for creating these semantic classes. A threshold of 0.85, for example, means that the cosine similarity between two words needs to be $>=0.85$ for the two words to be grouped in the same class. Words that are not included in the embedding vocabulary are ignored.\footnote{Please not that in  our experiments, we use in-domain embeddings trained on the same data sets that we use for scaling. This means that our embedding vocabulary has a high coverage and only words below the minimum token count (set when training the embeddings) are ``out-of-vocabulary''.} We proceed in a greedy fashion: once we encounter a word with a similarity higher than the threshold, we add it to the semantic class and remove it from the list, so that each word is assigned to exactly one semantic class. Different thresholds have an impact on the vocabulary size, i.e., on the number of distinct word forms in the input (see the {\em voc.} column in Table \ref{tab:wordfish-embed}). 
When the grouping is done, we can simply count word frequencies for all words in the same group, so that similar words are aggregated into one count. We can then construct the document frequency matrix over groups of similar words and use it as input for Wordfish.

Table \ref{tab:wordfish-embed} shows results on the European Parliament data set for correlations with ground-truth positions on EU integration. 
We notice only a small improvement for each of the three different similarity thresholds (0.95, 0.85, 0.8) and results are still significantly lower than the ones for SemScale with averaged embeddings.  
Interestingly, while the different thresholds have a strong impact on the vocabulary size in the input, these changes do not seem to have any effect on the results.

To sum up, this experiment shows that while word embeddings are an important ingredient in SemScale and contribute to its success, we can not trivially obtain a similar effect for Wordfish by feeding it less sparse and semantic-aware input. This is consistent with our results from Section \ref{sub:dtr} (Table \ref{tbl:vectors}) where SemScale outperformed Wordfish even when using sparse {\sc tf-idf} vectors as input for text scaling.

\subsection{Weakly Supervised Scaling}
\label{sub:wss}

As a final validation experiment, we present a comparison of SemScale and the most widely adopted \textit{supervised} text scaling algorithm, Wordscores (see Section \ref{sec:previous}). 
As SemScale was primarily designed for fully unsupervised scaling settings in which we assume that no ground truth positions are available for any of the documents, we first have to adjust the algorithm to be able to exploit this additional information. Our supervised extension of SemScale requires position scores for the two reference texts that represent extremes of the scale. For clarity, we refer to this (weakly) supervised extension of the original SemScale algorithm as \textit{SemScores}. 

In order to compare SemScores to Wordscores, we provide the same amount of supervision to both: we provide only the two documents that are the extremes (i.e., on each end) of the scale as reference texts and then evaluate the quality of the scalings, using the same correlation metrics as before.

\begin{table}[t]
\centering
\begin{tabular}{l ccc ccc }
\toprule
 & \multicolumn{3}{c}{5th Leg} & \multicolumn{3}{c}{6th Leg} \\
\cmidrule(lr){2-4}
\cmidrule(lr){5-7}

& PA & $r_P$ & $r_S$ & PA & $r_P$ & $r_S$ \\
\midrule

Wordscores  & 0.63 & 0.55 & 0.36 & \textbf{0.66} & \textbf{0.59} & \textbf{0.46} \\
&  (0.01) & (0.01) & (0.01) & (0.02) & (0.02) & (0.03) \\  

Semscores & \textbf{0.70} & \textbf{0.58} & \textbf{0.56} & 0.65 & 0.47 &	0.43 \\
& (0.01) & (0.0) & (0.03) & (0.01) & (0.00) & (0.01) \\
\bottomrule
\end{tabular}
\caption{Correlations between automatically produced positions with (a) Wordscores and (b) SemScores in a weakly supervised setting (with two reference texts denoting the opposite ends of the scale) and ground truth positions on European integration (Chapel Hill). Input: entire party texts concatenated from EP speeches. Results are averaged across different languages under study (standard errors are in brackets).\newline}
\label{tbl:supervised-euparl}
\end{table}

\begin{table}[t]
\centering
\begin{tabular}{l ccc ccc}
\toprule
 & \multicolumn{3}{c}{Single Election} & \multicolumn{3}{c}{Multiple Elections} \\
\cmidrule(lr){2-4}
\cmidrule(lr){5-7}

& PA & $r_P$ & $r_S$ & PA & $r_P$ & $r_S$ \\ \midrule
Wordscores  & \textbf{0.70} & 0.77 & 0.49 & \textbf{0.66} & \textbf{0.57} & \textbf{0.42} \\
&  (0.05) & (0.05) & (0.12) & (0.06) & (0.07) & (0.18) \\  

SemScores  & \textbf{0.70} & \textbf{0.80} & \textbf{0.53} & 0.63 & 0.46 & 0.33 \\
&  (0.06) & (0.02) & (0.12) & (0.06) & (0.07) & (0.16) \\ 
\bottomrule
\end{tabular}
\caption{Correlations between automatically produced positions with (a) Wordscores and (b) SemScores in a weakly supervised setting (with two reference texts denoting the opposite ends of the scale) and ground truth positions on Right-Left ideology (RILE, Manifesto Project). Input: entire party manifestos. Results are shown for both single election scaling and multiple election scaling settings (standard errors across languages are given in brackets).\newline}
\label{tbl:supervised-manifestos}
\end{table}

Results are shown in Table \ref{tbl:supervised-euparl}. 
We note that both algorithms produce scalings whose correlations with the Chapel-Hill EU integration positions 
drastically outperform the fully unsupervised scaling algorithms, Wordfish and SemScale (cf.~Table \ref{tbl:eu-int}). 
We can see similar improvements for scaling party manifestos where we compare correlations with RILE Right-Left ideological positions (see the comparison between Table \ref{tbl:supervised-manifestos} and Table \ref{tbl:manifestosbycountry}). 
This emphasises the benefits gained from minimal supervision in the form of positions of two documents at the opposite ends of the spectrum.

\section{Conclusion}

Years of research in text scaling have highlighted the fact that bag of words representations of documents, such as the ones employed by Wordfish, have the ability of capturing an underlying dimension across the collection under study, which often correlates with ideological positioning or attitudes towards a relevant topic, for instance, the European integration process. However, while such a scaling approach has supported a large number of different studies, it is inherently limited by the fact that it works at word-frequency level and does not consider any semantic representation of the text. In contrast to this, in this work we present SemScale, a new semantically-aware scaling method that exploits distributional semantic representations of the texts under study. We have provided empirical evidence that in many settings, by employing semantic information, scaling algorithms are able to better capture the European integration dimension, using the speeches from the European parliament as textual input. Moreover, we have shown how controlling for specific lexical and semantic information may lead to more robust position predictions, while at the same time facilitating the interpretability of produced positions for political scientists by reducing the size of the vocabulary under study (for instance when considering only nouns or only named entities instead of all words). 
We have also evaluated the newly introduced approach for directly inducing semantic representations of parties -- the so-called party embeddings -- in the context of scaling party manifestos, by coupling them with our graph-based scaling algorithm SemScale. Finally, we have presented a mechanism for extending SemScale to supervised scaling settings in which positions in the dimension of interest are available for some of the texts, demonstrating that even a small amount of supervision can massively improve the quality of an automatically induced scaling.  

While the results presented in this work seem promising, we believe that it is essential that our findings are treated with a healthy dose of scepticism, concretely, that the community (1) investigates the applicability and usefulness of semantic text scaling in a much wider set of scenarios and use cases and (2) bears in mind the limitations, some of which we have identified and analysed in this work (e.g., that semantic scaling is not particularly suitable when the texts span a long period of time, due to semantic and topical drift). To this end, we release together with this article the entire evaluation setting employed in our work and a Python implementation of SemScale (usable as a command-line tool). We hope that this effort will catalyse research on semantic text scaling and discovery of further settings in which it can support quantitative analyses in political science and its text-as-data community.

\begin{acks} 
This work is supported by the \grantsponsor{DFG}{German Research Foundation (DFG)}{https://www.dfg.de} under Grant No.~\grantnum[]{DFG}{139943784}. The authors are affiliated with the Collaborative Research Center SFB 884 ``Political Economy of Reforms'' and the Data and Web Science Group, University of Mannheim. Support for this research was provided by the German Research Foundation (SFB 884), projects C4 and B6. We thank Thomas Gschwend for comments on an earlier draft, Julian Bernauer and Konstantin Gavras for feedback on the implementation and Ashrakat Elshehawy for proofreading this paper. To view supplementary material for this article, please visit \url{https://federiconanni.com/semantic-scaling/}.
\end{acks}

%

\bibliographystyle{ACM-Reference-Format}
\bibliography{references}

\end{document}